\documentclass{llncs}
\usepackage{float}
\usepackage[T1]{fontenc}
\usepackage{amsmath}
\usepackage{comment}
\usepackage{longtable}

\usepackage{graphicx}
\usepackage{subfig}
\usepackage{hyperref}
\usepackage{xcolor}

\begin{document}
\title{Multimodal Representations for Teacher-Guided Compositional Visual Reasoning}
\titlerunning{Multimodal Teacher-Guided Compositional Visual Reasoning}
%
\author{Wafa Aissa \and
Marin Ferecatu \and
Michel Crucianu}

%
\authorrunning{[Aissa, W et al.]}
%
\institute{Cedric laboratory, Conservatoire National des Arts et Metiers, Paris, France
\email{firstName.lastName@lecnam.net}}
\maketitle              

\begin{abstract}
Neural Module Networks (NMN) are a compelling method for visual question answering, enabling the translation of a question into a program consisting of a series of reasoning sub-tasks that are sequentially executed on the image to produce an answer. NMNs provide enhanced explainability compared to integrated models, allowing for a better understanding of the underlying reasoning process. To improve the effectiveness of NMNs we propose to exploit features obtained by a large-scale cross-modal encoder. 
Also, the current training approach of NMNs relies on the propagation of module outputs to subsequent modules, leading to the accumulation of prediction errors and the generation of false answers. To mitigate this, we introduce an NMN learning strategy involving scheduled teacher guidance. Initially, the model is fully guided by the ground-truth intermediate outputs, but gradually transitions to an autonomous behavior as training progresses. This reduces error accumulation, thus improving training efficiency and final performance.
We demonstrate that by incorporating cross-modal features and employing more effective training techniques for NMN, we achieve a favorable balance between performance and transparency in the reasoning process.
\end{abstract}


\section{Introduction}\label{sec:introduction}
Visual reasoning, the ability to reason about the visual world, encompasses various canonical sub-tasks such as object and attribute categorization, object and relationship detection, comparison, and spatial reasoning. Solving this complex task requires robust computational models that can effectively capture visual cues and perform intricate reasoning operations. In recent years, deep learning approaches have gained prominence in tackling visual reasoning challenges, with the emergence of foundation models playing a key role in advancing the field.
Among the deep learning techniques employed for visual reasoning tasks, integrated attention networks, such as transformers \cite{Transformer}, have demonstrated remarkable success in natural language processing and computer vision applications, including image classification and object detection. These models leverage attention mechanisms to capture long-range dependencies and contextual relationships for language and vision inputs. However, despite their impressive performance, reasoning solely based on integrated attention networks may be susceptible to taking ``shortcuts'' and relying heavily on dataset bias. There is an increasing need to address the issue of interpretability and explainability, enabling reaserchers to understand the reasoning behind the model's predictions.
By leveraging transformer models as a backbone for VQA systems to encode language and visual information, we can adopt modular approaches that enable an improved understanding and transparency in the visual reasoning process. 
Modular approaches, such as neural module networks (NMNs), break down the problem (question) into smaller sub-tasks which can be independently solved and then combined to produce the final answer. The modular design confers the advantage of greater transparency and interpretability, as the model explicitly represents the various sub-tasks and their interrelationships.

In this paper we aim to bridge the gap between accuracy and explainability in visual question answering systems by providing insights into the reasoning process. To accomplish this, we propose an enhanced training approach for our modular network using a teacher forcing training technique \cite{TF_old}, where the ground truth output of an intermediate module is used to guide the learning process of subsequent modules during training. By employing this approach, our module gains the ability to learn its reasoning sub-task both in a stand-alone and end-to-end manners leading to improved training efficiency.

To evaluate the effectiveness of our approach, we conduct experiments using training programs sourced from the GQA dataset. This dataset provides a diverse range of scenarios, enabling us to thoroughly assess the capability of our approach to reason about the visual world. Through  experimentation and analysis, our results demonstrate the effectiveness of our proposed method in achieving explainability in VQA systems while maintaining a high degree of effectiveness.

In summary, this work makes two key contributions: first, the utilization of decaying teacher forcing during training, which enhances generalization capabilities, and second, the incorporation of cross-modal language and vision features to capture intricate relationships between text and images, resulting in more accurate and interpretable results.

The remaining sections of the paper are structured as follows: Sec.~\ref{sec:related-work} provides a discussion on related work, Sec.~\ref{sec:cross-modal-nmn} presents our cross-modal neural module network framework, and Sec.~\ref{sec:teacher-guidance} introduces our teacher guidance procedure. In Sec.~\ref{sec:experimental-protocol}, we outline the validation protocol, followed by the presentation of experimental results in Sec.~\ref{sec:results}. Finally, in Sec.~\ref{sec:conclusion}, we conclude the paper by synthesizing our findings and discuss potential future developments.

\section{Related work}\label{sec:related-work}

In this section, we begin by examining integrated and modular approaches employed in visual reasoning tasks. We then introduce the teacher forcing training method and its application to modular neural networks.

\textbf{Transformer networks.}
Transformers~\cite{Transformer} have been widely applied as foundation models for various language and vision tasks due to their remarkable performance. They have also been adapted for reasoning problems like Visual Question Answering (VQA). Notably, models such as ViLBERT~\cite{lu2019vilbert}, VisualBERT~\cite{li2019visualbert} and LXMERT~\cite{lxmert} have demonstrated interesting performance on popular VQA datasets like VQA2.0~\cite{balanced_vqa_v2} and GQA~\cite{gqa}. These frameworks follow a two-step approach: first, they extract textual and image features. Word embeddings are obtained using a pre-trained BERT~\cite{bert} model, while Faster RCNN generates image region bounding boxes along with their corresponding visual features. Subsequently, a cross-attention mechanism is employed to align the word embeddings with the image features, leveraging training on a diverse range of multi-modal tasks.

Despite the benefits of the integrated approaches, these models also have notable drawbacks. One prominent limitation is their lack of interpretability, making it challenging to understand---and debug, when necessary---the underlying reasoning process. Moreover, these models often rely on ``shortcuts'' in the reasoning, which means learning biases present in the training data. Consequently, their performance tends to suffer when confronted with out-of-distribution data, as shown on GQA-OOD~\cite{Kervadec_2021_CVPR}. This research also emphasizes the importance of employing high-quality input representations for the transformer model.

To address interpretability concerns, we use features produced by an off-the-shelf cross-modal transformer encoder in a step-by-step explainable reasoning architecture. This approach balances the power of the transformer model in capturing relationships between modalities with the ability to understand the reasoning process.

\textbf{Neural module networks.}
To enhance the transparency and emulate a human-like reasoning, compositional Neural Module Networks (NMNs) such as those introduced by~\cite{LearnReason} and~\cite{pvr} break down complex reasoning tasks into more manageable subtasks through a multi-hop reasoning approach. A typical NMN comprises a generator and an executor. The generator maps a given question to a sequence of reasoning instructions, known as a program. Subsequently, the executor assigns each sub-task from the program to a neural module and propagates the results to subsequent modules.

In a recent study by~\cite{MMN}, a meta-learning approach is adopted within the NMN framework to enhance the scalability and generalization capabilities of the resulting model. The generator decodes the question to generate a program, which is utilized to instantiate a meta-module. Visual features are extracted through a transformer-based visual encoder, while a cross-attention layer combines word embeddings and image features. Although the combination of a generator and an executor in NMNs may appear more intricate compared to an integrated model, the inherent transparency of the ``hardwired'' reasoning process in NMNs has the potential to mitigate certain reasoning ``shortcuts'' resulting from data bias.

A more recent study~\cite{me} has investigated the effects of curriculum learning techniques in the context of neural module networks. The research demonstrated that reorganizing the dataset to begin training with simpler programs and progressively increasing the difficulty by incorporating longer programs (based on the number of concepts involved in the program) facilitates faster convergence and promotes a more human-like reasoning process. This highlights the importance of curriculum learning in improving the training dynamics and enhancing the model's ability to reason and generalize effectively.

Interestingly, \cite{NEURIPS2021_9766527f} demonstrated that leveraging the programs generated from questions as additional supervision for the LXMERT integrated model led to a reduction in sample complexity and improved performance on the GQA-OOD (Out Of Distribution) dataset~\cite{Kervadec_2021_CVPR}. 

Building upon this, our work aims to capitalize on both the transparency offered by NMN architectures and the high-quality transformer-encoded representations by implementing a composable NMN that integrates multimodal vision and language features.

\textbf{Teacher forcing.}
Teacher forcing (TF)~\cite{TF_old} is a widely used technique in sequence prediction or generation tasks, especially in RNNs with an encoder-decoder architecture. It involves training the model using the true output as novel input, which helps improve prediction accuracy. However, during inference, the model relies on its own predictions without access to ground-truth information, leading to a discrepancy known as exposure bias.

Scheduled sampling (SS) is a notable approach to mitigating the train-test discrepancy in sequence generation tasks~\cite{SS}. It introduces randomness during training by choosing between using ground truth tokens or the model's predictions at each time step. This technique, initially developed for RNN architectures, has also been adapted for transformer networks~\cite{Transformer-SS}, aiding to align the model's performance during training and inference.

NMNs, on the other hand, are trained using only the output of a module as input for the next module, which has drawbacks. Errors made by an intermediate module can propagate to subsequent modules, leading to cumulative bad predictions. This effect is particularly prominent during the early stages of training when the model's predictions are close to random.

NMNs can leverage the TF strategy to enhance their training process. Initially, training begins with a fully guided schema, where the true previous outputs are used as input. As training progresses, the model gradually transitions to a less guided scheme, relying more on the generated outputs from previous steps as input. This gradual reduction in guidance and increased reliance on the model's own predictions, named decaying TF, helps NMNs better learn and adapt to the complexity of the task. With decaying TF, modules can conform to their expected behavior for their respective sub-tasks.

\section{Cross-modal neural module network}\label{sec:cross-modal-nmn}

\begin{figure}[t!] 

\includegraphics[width=\textwidth]{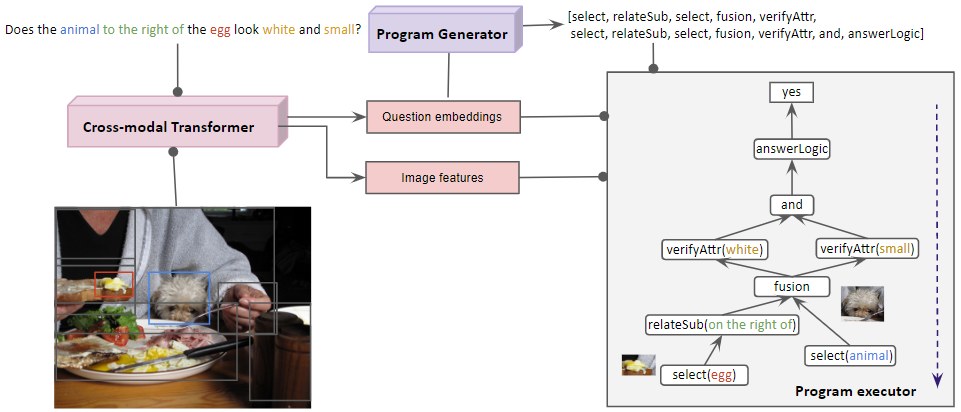}
\caption{The proposed modular VQA framework. Plain arrows represent the output flow, while dotted arrows represent the Multi-Task loss backward flow.
} \label{fig:model}
\end{figure}

Our model takes an image, question, and program triplet as input and predicts an answer. We extract aligned language and vision features for the image and question using a cross-modal transformer. The program, represented as a sequence of modules, is used to build an NMN, which is then executed on the image to answer the question (refer to Fig.~\ref{fig:model}). In the next subsections we detail the feature extraction process and describe the program executor.

\textbf{Cross-modal features.}
Compositional visual reasoning involves the ability to make logical and geometric inferences on complex scenes by leveraging both visual and textual information. This requires accurate representations of objects and questions. To address this, we employ LXMERT~\cite{lxmert}, a pretrained transformer model specifically designed for multi-modal tasks. LXMERT has demonstrated impressive performance across various tasks and serves as our feature extractor. In our approach, we discard the answer classification component and freeze the model's weights.
To extract cross-modal representations, we process the image~$I$ through the object-relationship encoder and the question~$Q$ through the language encoder. Then, the Cross-Modality Encoder aligns these representations and produces object bounding box features $v_j$ for each object $b_j$ in $I$, as well as word embeddings $txt_i$ for each word $w_i$ of $Q$.

\textbf{Neural modules.} Our NMN approach tackles complex reasoning tasks by decomposing them into simpler sub-tasks, inspired by human reasoning skills like object detection, attribute identification, object relation recognition, and object comparison.
We developed a library of modules tailored to address specific sub-tasks. These modules are designed to be intuitive and interpretable, using simple building blocks like dot products and MLPs. They are categorized into three groups: attention, boolean, and answer modules. For instance, the \texttt{Select} attention module focuses on detecting object bounding boxes by applying an attention vector to the available bounding boxes within an image. On the other hand, boolean modules like \texttt{And} or \texttt{Or} make logical inferences, while answer modules such as \texttt{QueryName} provide probability distributions over the vocabulary of possible answers. To get a glimpse of the variety within our module library you can refer to Table~\ref{modules_small}, which showcases an example from each module category. 

\begin{table}[h!]
\centering
\caption{Sample module definitions. $S$:~softmax, $\sigma$:~sigmoid, $r$:~RELU, $\mathbf{W}_i$:~weight matrix, $\mathbf{a}$:~attention vector ($36\times1$), $\mathbf{V}$:~visual features ($768\times36$), $\mathbf{t}$:~text features ($768\times1$),
$\odot$: Hadamard product.} 
\label{modules_small}
{\small
\centering
\setlength\tabcolsep{1.5pt}
   \begin{tabular}{ | c | c | c | c | }
     \hline
      Name & Dependencies & Output & Definition  \\ \hline
      Select & $-$ & attention & $\begin{matrix}  
      \mathbf{x} = r(\mathbf{W}\,\mathbf{t}),  
      \mathbf{Y} = r(\mathbf{W}\mathbf{V}) ,
      \mathbf{o} = S(\mathbf{W}(\mathbf{Y}^T \mathbf{x})) \\ \end{matrix}$ \\ \hline 
    RelateSub & [$\mathbf{a}$] & attention & 
     $\begin{matrix} 
    \mathbf{x} = r(\mathbf{W}\,\mathbf{t}),
     \mathbf{Y} = r(\mathbf{W}\mathbf{V}) ,
     \mathbf{z} = S(\mathbf{W}(\mathbf{Y}^T \mathbf{x})) \\
     \mathbf{o} = S(\mathbf{W}(\mathbf{x} \odot \mathbf{y} \odot \mathbf{z}))
    \end{matrix}$ \\ \hline
      VerifyAttr & [$\mathbf{a}$] & boolean & 
     $\begin{matrix} 
     \mathbf{x} = r(\mathbf{W}\,\mathbf{t}) ,
     \mathbf{y} = r(\mathbf{W}(\mathbf{V}\,\mathbf{a}) ,
     \mathbf{o} = \sigma(\mathbf{W}(\mathbf{x} \odot \mathbf{y} ))
     \end{matrix}$\\
     \hline
      And & [$\mathbf{b}_1$,$\mathbf{b}_2$] & boolean & $\begin{matrix} \mathbf{o} = \mathbf{b}_1 \times \mathbf{b}_2 \\ \end{matrix}$ \\ \hline 
      ChooseAttr & [$\mathbf{a}$] & answer & $\begin{matrix} 
     \mathbf{x} = r(\mathbf{W}\,\mathbf{t}) ,
     \mathbf{y} = r(\mathbf{W}(\mathbf{V}\,\mathbf{a}) ,
     \mathbf{o} = S(\mathbf{W}(\mathbf{x} \odot \mathbf{y} ))
     \end{matrix}$ \\ \hline
      QueryName & [$\mathbf{a}$] & answer &  $\begin{matrix} 
      \mathbf{y} = r(\mathbf{W} (\mathbf{V}\,\mathbf{a})) , 
      \mathbf{o} = S(\mathbf{W}\,\mathbf{y})
      \end{matrix}$\\ \hline
   \end{tabular}
   }
\end{table}

\textbf{Modular network instantiation.} 
A program in our framework consists of a sequence of neural modules (Table~\ref{modules_small}). These modules are instantiated within a larger Neural Module Network (NMN) following the program sequence. Each module has dependencies, denoted as $d_m$, which allow it to access information from the previous modules, and arguments, denoted as $txt_m$, which condition its behavior.
For example, the \texttt{FilterAttribute} module, which relies on the output of the \texttt{Select} module, aims to shift attention to the selected objects by considering the attribute that corresponds to the provided text argument.
To handle module dependencies, the program executor employs a memory buffer to store the outputs, further used as inputs for subsequent modules. This approach also enables the computation of multi-task losses (see Sec.~\ref{sec:teacher-guidance}) by comparing the outputs produced by the modules with the expected ground-truth outputs.

\section{Teacher guidance for neural module networks}
\label{sec:teacher-guidance}

\begin{figure}[t!]
\includegraphics[width=\textwidth]{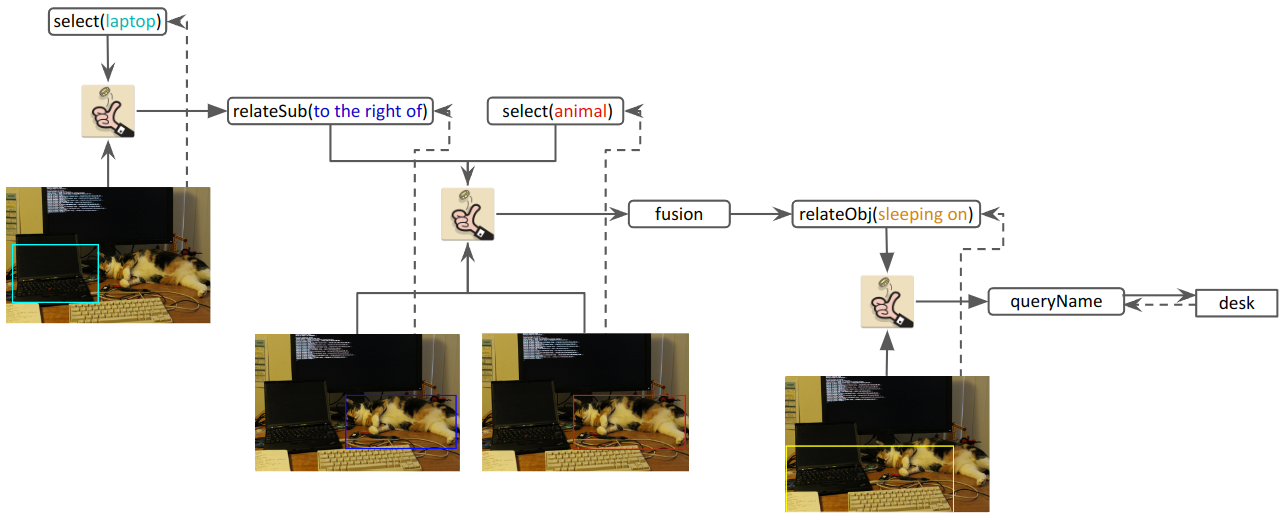}
\caption{The teacher guidance for the program execution process related to the question `On what is the animal to the right of the laptop sleeping?'. Plain arrows represent input guidance and dotted arrows represent the output feedback.} \label{fig:TF-for-NMN}
\end{figure}

To achieve explainable reasoning, we use teacher forcing (TF) to guide the modules by providing them with ground-truth inputs. We also employ a multi-task (MT) loss to provide feedback and correct their behaviors towards the expected intermediate outputs. This process is illustrated in Fig.~\ref{fig:TF-for-NMN}.

Given a program $p$, a question $q$ and image $I$, the modular network executes $p$ on $I$ with the textual arguments $txt_m$ encoded in $q$, producing an answer $a$. This can be represented as $a = p(I, q)$, where $ p = m_1 \circ m_2 \circ ... \circ m_n $ denotes the sequential execution of $n$ modules within the program. Each module $m_t$ inputs the output of the previous module $m_{t-1}$  and performs a specific computation or reasoning step to contribute to the final answer. The NMN is trained by minimizing the cross entropy loss $L_{CE}$ over the set of $(p, q, I, a)$ examples. In fact, when a module is provided with its golden input and expected output, it is independently optimized to perform its specific sub-task. However, when modules are jointly trained in a sequential manner, they learn to adapt their behaviors to work together and engage in explicit reasoning without taking shortcuts. This collaborative approach enables the modules to develop a deeper understanding of the task and enables them to perform complex reasoning operations.

From a back-propagation perspective, during the early stages of training, the gradients are computed based on the losses of individual modules when processing correct inputs. As a result, the backward gradient flow of the MT loss is interrupted at the first ground truth input.
However, in the case of collaborative module interactions without TF, the full back-propagation can be computed. 
The intermediate outputs are preserved in continuous form throughout the program execution, enabling the flow of backward gradients between modules. Errors and updates can be propagated through the entire network, facilitating effective learning and enhancing the overall performance of the NMN. We give details about our guidance mechanism in the following subsections.

\textbf{Input guidance.} The modules receive input guidance through decaying teacher forcing (TF). As shown in Fig.~\ref{fig:TF-for-NMN}, at each reasoning step $t$ the executor randomly decides whether to use the predicted output $\hat{o}_{t-1}$ or the ground-truth output $o^*_{t-1}$ from the previous module $m_{t-1}$ as its input. This decision is made by flipping a coin, where $o^*_{t-1}$ is chosen with a probability of $\epsilon_e$ and $\hat{o}_{t-1}$ is chosen with a probability of $1-\epsilon_e$.
The coin-flipping process for input selection occurs at each reasoning step, allowing the model to train on various sub-programs. The probability $\epsilon_e$ of selecting $o^*_{t-1}$ depends on the epoch number $e$. As training progresses and the epoch number increases, $\epsilon_e$ decreases, giving more preference to the module's predictions over the ground-truth intermediate outputs.

\textbf{Output feedback.} We employ a multi-task (MT) loss approach to provide feedback to the modules based on their outputs. The loss consists of a weighted sum $L = \alpha L_{att} + \beta L_{bool} + \gamma L_{answer}$ of individual losses for the attention modules, boolean modules and answer modules, with $\alpha$, $\beta$ and $\gamma$ scaling factors.
Each module is assigned its own average loss, considering its frequency of appearance, to prevent overemphasis on frequent modules at the expense of infrequent ones. 

For Boolean modules, we rely on the provided answer to infer the module's output and generate the intermediate Boolean outputs. However, for attention modules, it is necessary to establish correspondences between the bounding boxes in the image graph and those obtained from Faster-RCNN, which brings us to the issue of the ground-truth intermediate outputs, detailed in the following.

\textbf{Soft matching and hard matching.} Two mapping techniques, namely hard matching and soft matching, are employed to align the ground-truth bounding boxes with those obtained from the feature extractor. In the hard matching approach, a ground-truth bounding box $b_g$ is matched with the bounding box $o^*_i$ from the feature extractor that has the highest Intersection over Union (IoU) factor. On the other hand, the soft mapping matches $b_g$ with all $o^*_i$ that have an IoU value above a threshold, resembling a multi-label classification task.
The choice of the matching technique directly affects the representation of the attention intermediate output vectors. Hard mapping produces one-hot-like vectors, while soft mapping multi-label vectors, with one(s) for positive matching and zeros for negative matching boxes.
It is important to acknowledge that not all modules have ground-truth outputs that can be extracted.

\section{Protocol design}\label{sec:experimental-protocol}

\textbf{Dataset \& metrics.}
The GQA balanced dataset~\cite{gqa} consists of over 1 million compositional questions and 113,000 real-world images. The questions are represented by functional programs that capture the reasoning steps involved in answering them. 
To ensure consistent evaluation, the dataset authors suggest using the \texttt{testdev} split instead of the \texttt{val} split when utilizing object-based features due to potential overlap in training images. In line with the latter and following LXMERT, our model is trained on the combined \texttt{train+val} set. For testing, we evaluate the model's performance on the \texttt{testdev-all} split from the unbalanced set. This allows us to gather additional examples and gain a comprehensive understanding of the NMN's behavior.
To simplify the module structure in the GQA dataset, we consolidate specific modules into more general ones based on similar operations. For example, modules like \texttt{ChooseHealthier} and \texttt{ChooseOlder} are combined into \texttt{ChooseAttribute} module, with an argument $txt_m$ specifying the attribute to select. This reduces the number of modules from 124 to 32. 
Our experiments directly utilize the pre-processed GQA dataset programs, with a specific focus on evaluating the teacher forcing training on the Program Executor module. While our system employs a transformer model as a generator to convert the question into its corresponding program, this task is relatively straightforward compared to the training of the executor. As in previous studies ~\cite{pvr,MMN}, we achieve nearly perfect translation results on \texttt{testdev-all}.

We assess the performance of our approach by measuring answer accuracy. Additionally, we conduct a qualitative evaluation of the intermediate outputs, visualized through plotted images in Sec.~\ref{sec:results}.

\textbf{Evaluated methods.}
As presented in the previous sections, we propose two contributions to improve neural module networks for VQA. First, we use teacher guidance during training, leading to better generalization. Second, we leverage cross-modal language and vision features to capture complex relationships between text and images, resulting in more accurate and interpretable results.

We use the following notations to describe the various experimental setups: \\
- \textbf{LXV}: Employ the cross-modal representations from the LXMERT model~\cite{lxmert}. \\
- \textbf{TF}: Apply decaying teacher forcing to guide the inputs of the modules. \\
- \textbf{MT}: Apply multi-task losses to guide the expected outputs of the modules. \\
- \textbf{Soft}: Use the soft matching technique described in Sec.~\ref{sec:teacher-guidance}. \\
- \textbf{Hard}: Employ the hard matching technique described in Sec.~\ref{sec:teacher-guidance}. \\
- \textbf{BertV}: Use unimodal contextual language and vision representions, where contextual text embeddings are extracted by the BERT model~\cite{bert} and Faster-RCNN bounding boxes features are provided by the GQA dataset~\cite{gqa}. \\
- \textbf{FasttextV}: Employ unimodal non-contextual fastText embeddings~\cite{fasttext} along with Faster-RCNN bounding boxes features.

\section{Results analysis}\label{sec:results}

In our evaluation, we begin by comparing the different teacher-guided training strategies. We also compare the impact of the addition of the multi-task losses and its correlation with the decaying TF along with the soft and hard matching techniques. 
Later, we compare the usage of multi-modal representations against uni-modal representations. 

\textbf{Analysis of the teacher guided training.}
We aim to enable modular reasoning for visual question answering on the GQA dataset. We evaluate the effectiveness of our approach by measuring the answer accuracy of several models (described in Sec.~\ref{sec:experimental-protocol}), and report the results in Table.~\ref{tab:TF}. Overall, our findings demonstrate that using a combination of input guidance (denoted as \textbf{TF}) and output guidance (\textbf{MT}) achieves the highest accuracy, with a score of 63.2\%.

\begin{table}[b!]
\caption{Performance of various training methods on the \texttt{testdev-all} set.}\label{tab:TF}
\centering
\begin{tabular}{|l|l|}
\hline
Model & accuracy  \\ 
\hline
LXV-TF-hard &   {0.548}   \\ 
LXV-MT-hard &   {0.598} \\ 
LXV-TF-MT-hard  & {0.630}  \\ 
LXV-TF-soft &   {0.536}  \\ 
LXV-MT-soft &   {0.563}  \\ 
LXV-TF-MT-soft  & {\textbf{0.632}}  \\ 
\hline
\end{tabular}
\end{table}

When comparing \textbf{LXV-TF} (decaying teacher forcing) with \textbf{LXV-MT} (multi-task loss), we observe that the multi-task loss alone achieves higher accuracy than using decaying teacher forcing alone. This can be attributed to the fact that when using TF alone, the final loss $L$ is solely determined by the answer modules loss $L_{answer}$ and during early training stages, the application of TF limits the backpropagation process, preventing it from reaching the first modules of the programs. As a result, the impact of $L_{\text{answer}}$ on initial modules is limited.

Interestingly, the combination of multi-task loss and decaying teacher forcing exhibits complementary effects, leveraging the strengths of both techniques to enhance training dynamics and overall performance.

To assess the effectiveness of the decaying teacher forcing guidance, we compare \textbf{LXV-MT} against \textbf{LXV-TF-MT}. The \textbf{TF} guidance has led to accuracy improvements in both \textbf{soft} and \textbf{hard} matching settings for NMN. 
This technique can be viewed as a form of curriculum learning, where the model trains on programs of increasing length and complexity. During training, we observed a faster increase in accuracy for the models using \textbf{TF} compared to those without TF, as the answer modules receive ground truth inputs in the early stages. As training progresses, the training performance continues to improve until it reaches a peak, after which it slightly degrades due to the reduced use of TF and the modules adjusting to collaborative functioning. Nonetheless, as training continues, the testing performance surpasses that of the models without TF.

When combining the \textbf{MT} loss with \textbf{LXV-TF}, modules are optimized based on their intermediate outputs losses and they can benefit from the additional guidance provided by the back-propagation of $L_{att}$ and $L_{bool}$.   We reach the best performances outlined by \textbf{LXV-TF-MT-soft} and \textbf{LXV-TF-MT-hard}. The increase in accuracy ranges from $+8.2\%$ in the \textbf{hard} matching setting to $+9.6\%$ in the \textbf{soft} matching setting.

\textbf{Unimodal \emph{vs} cross-modal representations.}
We measure the impact of different input representations on the performance (see Table.~\ref{tab:unimodal-vs-cross-modal}). For unimodal embeddings we encode the question with fastText word embeddings or BERT language model, and the image with Faster-RCNN features. For cross-modal representations, we encode the question and the image with LXMERT.
The experiments are conducted using our best training strategies from the previous section, i.e. we employ the TF guidance and the MT loss for all the experiments.

When comparing \textbf{fastText} and \textbf{BERT}, empirical observations indicate that \textbf{BERT} tends to achieve better performance when utilizing hard matching, which involves a focused and selective attention mechanism. Conversely, \textbf{fastText} demonstrates improved performance with the soft matching mechanism, enabling a multi-label approach. The choice between these matching mechanisms relies on the inputs of the models and the training strategy, as each model may demonstrate superior performance in different scenarios.  

\begin{table}[b!]
\centering
\caption{Language and vision representations results on \texttt{testdev-all}. 
\centering
}\label{tab:unimodal-vs-cross-modal}
\begin{tabular}{|l|l|}
\hline
Model & accuracy \\ 
\hline
FasttextV-TF-MT-hard &  {0.495}  \\ 
BertV-TF-MT-hard & {0.506}  \\ 
LXV-TF-MT-hard & {0.630}  \\ 
BertV-TF-MT-soft &  {0.485}  \\ 
FasttextV-TF-MT-soft &  {0.511}  \\ 
LXV-TF-MT-soft & \textbf{0.632}  \\ 
\hline
\end{tabular}
\end{table}

Cross-modal aligned features provided by LXMERT (denoted as~\textbf{LXV}) have shown a significant increase in accuracy, with a $+12.1\%$ improvement when using soft matching and a $+12.4\%$ improvement when using hard matching. This validates our intuition that leveraging cross-modal features pretrained on diverse tasks and large datasets can greatly benefit NMNs. By incorporating these features, the modular reasoning process is performed with a better understanding of word embeddings and bounding box features, leading to enhanced performance and more accurate predictions.

\begin{figure}[h!] 
\includegraphics[width=\textwidth]{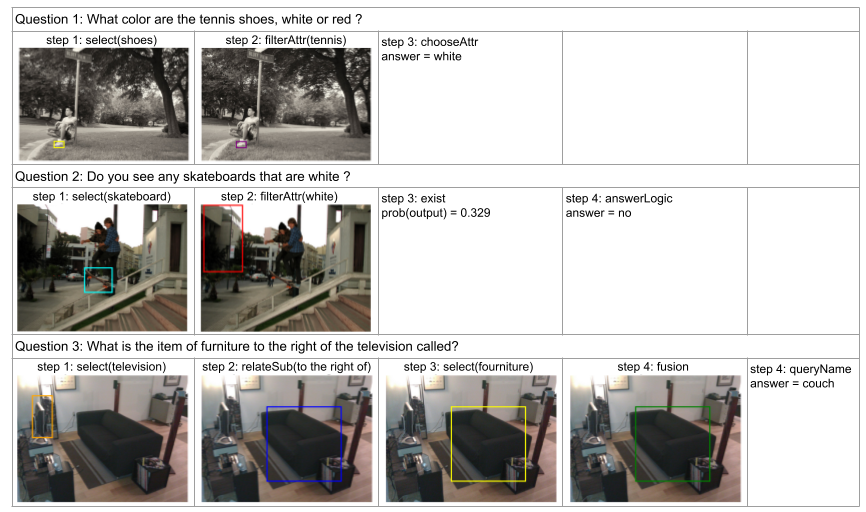}
\caption{Visualization of the reasoning process.
} \label{fig:outputs}

\end{figure}

\textbf{Qualitative analysis of the modular approach.}
In Fig. \ref{fig:outputs}, we illustrate the reasoning process for three different questions. We highlight the bounding boxes with the highest attention values from the attention output vector. For boolean modules, we display the output probability and finally the predicted answer.
Taking ``Question 2'' as an example, the first step successfully selects the skateboard as the object of focus. In the second step, the attention shifts to white objects. Since the skateboard is not white, the attention is then redirected to the white building. The ``exist'' module assesses if there is an object with a high attention value and produces a probability score based on which the answer is predicted.
These examples demonstrate the explainability of our approach and the ability to trace the model's decision-making process.

\vspace{-3ex}

\section{Conclusion} \label{sec:conclusion}

We have presented a neural module framework trained using a teacher guidance strategy, which has demonstrated several key contributions. First, our approach enhances generalization and promotes a transparent reasoning process, as evidenced by the experimental results on the GQA dataset. Additionally, the utilization of cross-modal language and vision features allows to capture intricate relationships between text and images, leading to improved accuracy. By harnessing our proposed approach, the neural modules acquire the capability to learn their reasoning sub-tasks both independently and in an end-to-end manner. This not only enhances training efficiency but also increases the interpretability of the system, allowing for a better understanding of the underlying reasoning processes.
In addition to the aforementioned contributions, our work paves the way to a better understanding of NMNs for the task of visual reasoning. Promising directions include extending our approach to other visual reasoning datasets for a broader evaluation, and exploring alternative training strategies to enhance performance and efficiency.

\textbf{Acknowledgments.} We thank Souheil Hanoune for his insightful comments. This work was partly supported by the French Cifre fellowship 2018/1601 granted by ANRT, and by XXII Group.

\bibliographystyle{splncs04}
\bibliography{mybibliography}

\end{document}